\documentclass[a4paper,11pt]{article}

\usepackage{times}
\usepackage{courier}
\usepackage{blindtext}

\usepackage[english]{babel}
\usepackage[utf8]{inputenc}
\usepackage[T1]{fontenc}
\usepackage{ragged2e}
\usepackage{subcaption}
\usepackage{float}
\usepackage{makecell}
\usepackage{enumitem}
\usepackage{textcomp}
\usepackage{scalerel}
\usepackage{lscape}
\usepackage{svg}
\usepackage{lineno}

\usepackage{graphicx}

\usepackage{amssymb, amsmath} 

\usepackage{setspace}         

\usepackage{tabularx}

\usepackage{url}              

\usepackage{hyperref}
\hypersetup{
colorlinks=false,
pdfborder=0 0 0	
}

\usepackage[babel,german=quotes]{csquotes}

\usepackage[backend=bibtex8, sorting=none]{biblatex}
\bibliography{bibliography}

\begin{document}

\title{Automatic Bat Call Classification \\using Transformer Networks}
\author{Frank Fundel, Daniel A. Braun, Sebastian Gottwald \\
Institute of Neural Information Processing, Ulm University}
\date{First submission on: March 22, 2023\\
Accepted on: September 1, 2023}

\maketitle

\section*{Abstract}
Automatically identifying bat species from their echolocation calls is a difficult but important task for monitoring bats and the ecosystem they live in. Major challenges in automatic bat call identification are high call variability, similarities between species, interfering calls and lack of annotated data. Many currently available models suffer from relatively poor performance on real-life data due to being trained on single call datasets and, moreover, are often too slow for real-time classification. Here, we propose a Transformer architecture for multi-label classification with potential applications in real-time classification scenarios. We train our model on synthetically generated multi-species recordings by merging multiple bats calls into a single recording with multiple simultaneous calls. Our approach achieves a single species accuracy of 88.92\% (F1-score of 84.23\%) and a multi species macro F1-score of 74.40\% on our test set. In comparison to three other tools on the independent and publicly available dataset ChiroVox, our model achieves at least 25.82\% better accuracy for single species classification and at least 6.9\% better macro F1-score for multi species classification.

\section*{Keywords}
computational bioacoustics, attention, Transformer, echolocation, species identification, acoustic monitoring, bat calls

\newpage

\section{Introduction}
Bats play a vital role in maintaining ecological balance in various ecosystems worldwide. They provide essential pest management for agricultural crops, act as primary predators of mosquitoes and other nocturnal flying insects, pollinate and disperse plant seeds, and even contribute to the formation of certain cave ecosystems through their guano \cite{neuweiler2000biology,Whybatsm71:online}. Moreover, bats serve as excellent indicators of biodiversity and environmental health \cite{Whybatsm71:online}. Monitoring bat populations is therefore crucial, particularly considering the decline of species, and some being on the verge of extinction, as observed in Germany, for instance \cite{skiba2003europäische,Säugetie62:online}. This task is, however, incredibly challenging, because bats only hunt at night, travel at high speeds, and are audibly silent for human observers, such that the only non-invasive method of monitoring bats is based on recording and categorizing their ultrasonic echolocation calls. As classifying hours of recordings manually is tedious, automatic detection and classification methods have been studied for many years.

Early methods used frequency analysis tools for feature extraction and decision trees for classification \cite{Herr, Adams}. Later, simple machine learning methods like Multi-Layer Perceptrons (MLPs) \cite{parsons, Britzke, Ayala-Berdon, PREATONI, ARMITAGE, NUNEZ, Jennings}, Linear Discriminant Analysis (LDA) \cite{Ayala-Berdon, Russo, PREATONI, ARMITAGE}, Support Vector Machines (SVMs) \cite{redgwell,ruiz, ARMITAGE, NUNEZ}, or ensembles of MLPs \cite{walters} were used for classifying up to 44 different pre-extracted features. More recent methods use simple Convolutional Neural Networks (ConvNets) \cite{batdetective, Zualkernan, Khalighifar, Alipek} and Residual Neural Networks (ResNets) \cite{schwab, tabak, CHEN} to detect and classify single calls, and also Recurrent Neural Networks (RNNs) to separate echolocation calls from social calls \cite{zhang}. This is in line with neighboring research fields, where ConvNets and RNNs are used to classify vocalizations of whales \cite{bermant, Bergler, Shiu} or birds \cite{emreçakır2017convolutional, MORALES2022101909, Adavanne, Grill, Dufourq, Sprengel}, for example.

\label{sentence:problems}
However, there are several challenges associated with automatic bat call classification. For instance, distinguishing between closely related bat species such as \emph{Myotis brandtii} and \emph{Myotis mystacinus} can be difficult due to their similar calls \cite{skiba2003europäische, schwab}. Additionally, bat call recordings exhibit significant variability, influenced by factors such as environmental conditions,  flying velocity \cite{neuweiler2000biology, BatsNotBirds} and particular acoustic behaviours such as social calls and feeding buzzes \cite{Prat2016}. Another issue is the limited availability of annotated data, as manual classification of bat calls requires expertise and extensive experience. Furthermore, since multiple bats of different species often call simultaneously, the presence of overlapping calls makes detection and classification challenging. These factors contribute to the overall difficulty in achieving accurate classification performance, resulting in poor generalization due to high variability in a relatively small amount of training data. Moreover, models trained on single, non-overlapping calls may struggle with overlapping calls encountered in real-world scenarios, resulting in subpar performance \cite{schwab}.


Most existing approaches in bat call classification primarily focus on individual calls and disregard the temporal succession within call sequences. However, classifying \textit{sequences} of bat calls could potentially address the high variability issue by capturing changes in calls over time, including transitions between different flight patterns and speeds. Additionally, analyzing sequence data may alleviate difficulties in classifying overlapping calls when training on data with interfering calls from different species, where current models struggle—refer to Figure \ref{tbl:comp_mixed}. Finally, the use of large models like ResNet-50 can have performance issues, particularly in terms of inference time, that typically increases with the number of model parameters. This becomes particularly impractical when classifying long recordings, especially when running on a CPU.

Here, we present BioAcoustic Transformer (BAT), a fast and light-weight end-to-end architecture for classifying overlapping multi-species bat call sequences. Our approach utilizes a small ConvNet-Transformer hybrid model that operates on spectrogram representations of bat call recordings. Unlike previous models that rely on detecting individual calls and classifying them separately, BAT is trained on synthetically generated multi-species call sequences. This approach allows us to perform multi-label classification, enabling the detection of different bat species within a single analysis. By leveraging this methodology, our model achieves improved efficiency and accuracy in handling overlapping calls.


\section{Methods}

\subsection{Data acquisition and preprocessing}
\label{sec:dataprep}
\begin{figure}[H]
    \centering
    \includegraphics[width=0.9\textwidth]{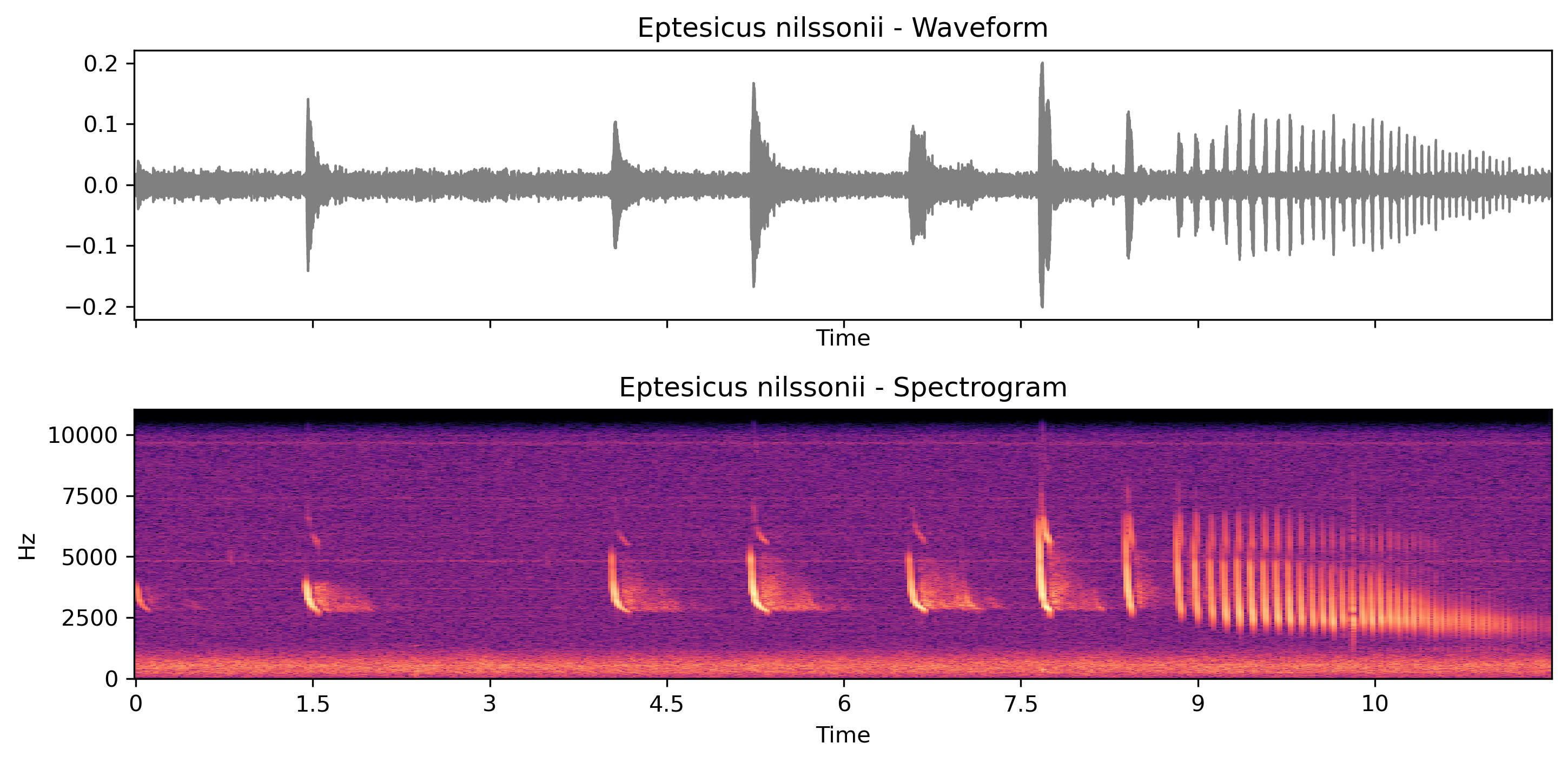}
    \caption{Exemplary recording of \emph{Eptesicus nilssonii} from the Skiba dataset.  The waveform shows distinct peaks indicating the occurrence of bat calls, alongside background noise. Specifically, the calls of \emph{Eptesicus nilssonii} exhibit significant frequency modulation, typically falling within the range of 24-27 kHz. Notably, the presence of a final buzz at the end of the call can be observed, representing the bat's approach towards its prey. It is important to note that due to the time expansion factor, the frequency values need to be multiplied by 10. Hence, 2,500 Hz corresponds to 25,000 Hz in the original recording.}
    \label{fig:dataset_example}
\end{figure}

\begin{figure}[H]
    \centering
    \includegraphics[width=\textwidth]{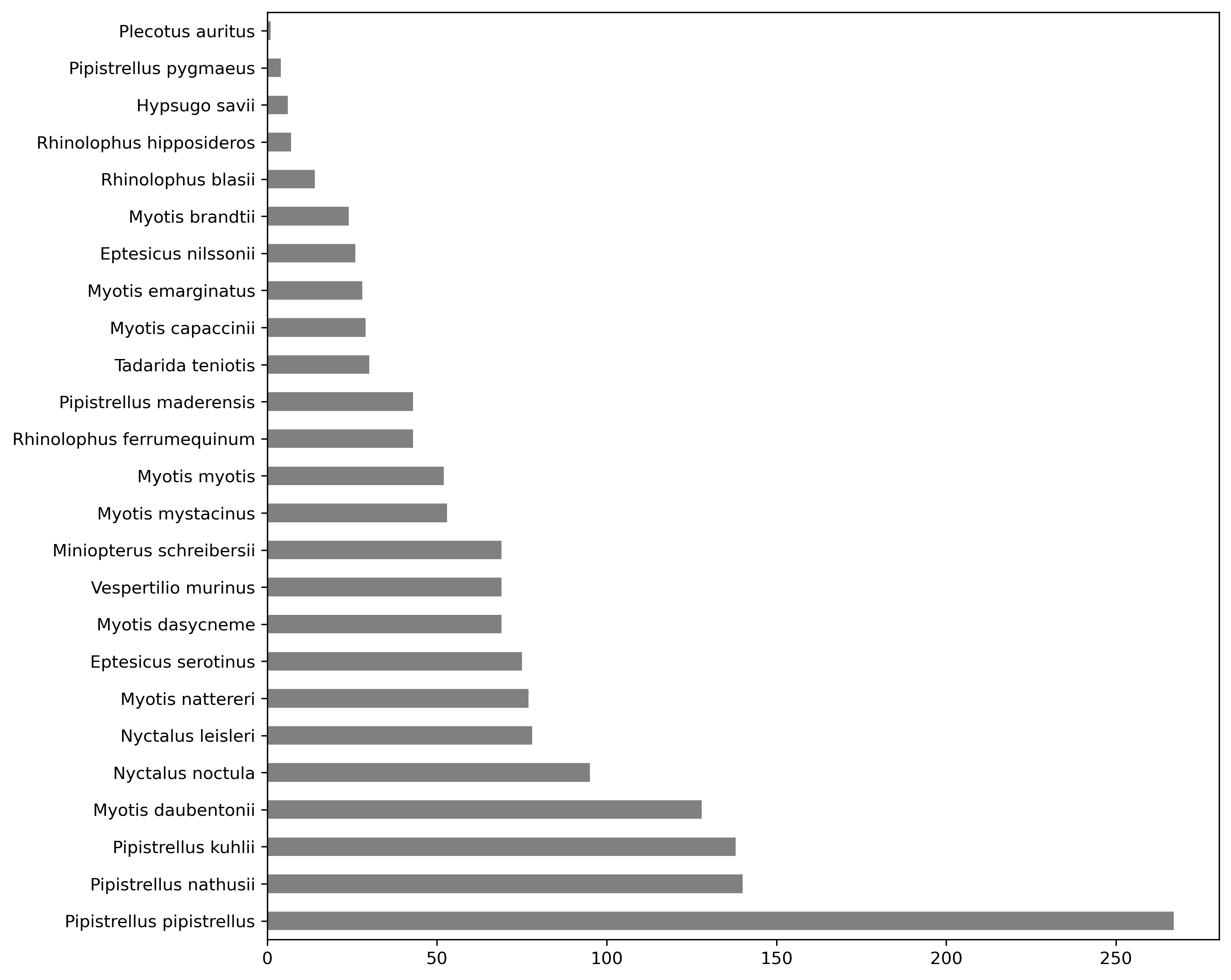}
    \caption{Histogram of recordings per species in the Skiba dataset.}
    \label{fig:dataset_distribution}
\end{figure}

Our dataset is based on the Skiba dataset \cite{skiba2003europäische} obtained from the \textit{Museum für Naturkunde Berlin}. This dataset comprises more than 1,500 recordings and over 45,000 individual calls, encompassing 29 bat species. The dataset consists of approximately 10 GB of WAV audio files. 
All full spectrum recordings in this dataset are based on the "Pettersson D980" device, with a consistent time expansion ratio of 1:10, a sample rate of 96,000, and a bit depth of 24. Each recording in the dataset has been classified by an expert, showcasing a high degree of variability and absence of overlap between calls—refer to Figure \ref{fig:dataset_example} for an illustrative example.  Counting the number of recordings per species reveals that there are less recordings of the rarer species---see Figure \ref{fig:dataset_distribution}. To supplement the training process, we aimed to incorporate a separate model trained on overlapping call sequences, which more closely resembles natural conditions. However, due to the lack of a sufficiently large existing dataset containing overlapping bat calls, we generated our own synthetic dataset of overlapping call sequences.


The recordings are prefiltered using a 10th-order butterworth high-pass filter, and downsampled from 96,000 samples per second to 22,050. We experienced no difference in model performance when using higher sample rates than 22,050 samples per second. We then split the recordings into randomized training (60\%,  11,323 recordings), test (25\%,  4,980 recordings) and validation (15\%,  2,891 recordings) sets and stored them in hdf5-files for faster loading (streaming). Importantly, we split the recordings before generating sequences, so sequences from the same recording never occur in two different sets. We primarily focus on German bat species in our analysis; however, some of them were significantly underrepresented in the dataset, so we made the decision to exclude them. As a result, the final dataset comprises 18 species.

Our dataset only contains recordings where a single bat is calling at any one time, each having a corresponding label. To simulate "mixed calling" scenarios, we synthesized these instances by combining multiple randomly sampled sequences and their corresponding one-hot encoded labels.
At any given time, between one to three single bat call sequences are randomly selected for mixing. The mixing process involves adding the time signals together and dividing the result by the number of mixed signals. Once mixed, the signals are transformed into their spectrogram representations. Subsequently, each spectrogram undergoes filtering to remove constant noise across each frequency band.
Due to independent random mixing for each batch, it is highly likely that each batch is unique. The same mixing approach was applied to the sequences used for validation and testing.

\subsection{Model architecture}
\label{sec:method}
Our BioAcoustic Transformer (BAT) is a ConvNet-Transformer hybrid model on spectrograms. Intuitively, the ConvNet extracts local spatial features of each time patch of the spectrogram and the Transformer detects global temporal features of the whole sequence. More precisely, the ConvNet is used to embed each patch in the time domain, where the patch size is chosen to have the average length of a single call. The subsequent attention mechanism in the Transformer can then correlate each embedded patch with every other embedded patch of the sequence. The possibility of such hybrid architectures were already mentioned by the authors of the original Transformer \cite{AIAYN}, anticipating that the linear embeddings to which subsequent self attention layers are applied in the original architecture can be replaced by various other embedding networks---see for example \cite{https://doi.org/10.48550/arxiv.2107.00781,https://doi.org/10.48550/arxiv.2207.03450}.

\begin{figure}[H]
    \centering
    \includegraphics[width=0.9\textwidth]{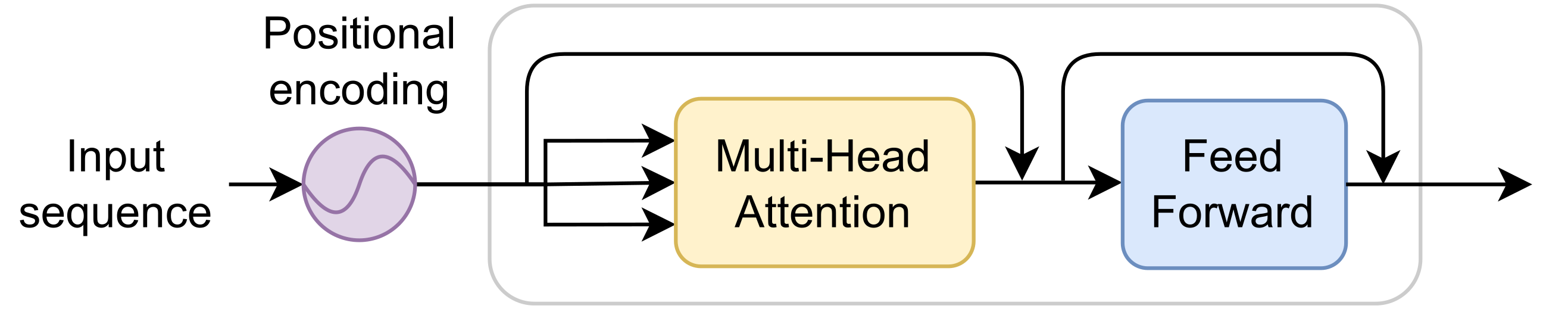}
    \caption{The Transformer encoder architecture.}
    \label{fig:Transformer}
\end{figure}

The Transformer architecture was first introduced in 2017 by Vaswani et al. \cite{AIAYN} in the context of language processing (NLP), in particular for translation tasks. Soon afterwards it was discovered that its base architecture (Transformer encoder block) is very versatile and nowadays almost every model that tops the state-of-the-art charts, especially in sequence processing tasks, contains a Transformer-like part somewhere in its architecture. A basic Transformer encoder block is displayed in Figure \ref{fig:Transformer}. First, the input sequence is embedded token-wise into a latent representation (one vector for each token), usually containing positional information, also known as \textit{positional encoding}. Every attention unit in the transformer determines for each token three vectors (Query vector, Key vector Value vector) that depend on the token itself and all the other tokens. From these vectors, attention weights can be calculated between all token pairs simultaneously. These attention weights are then used to produce an output that corresponds to a weighted sum of value vectors for each token. In order to consider multiple weighting schemes reflecting multiple relevance relationships,
there are typically multiple copies of attention heads with different \textit{Query-Key-Value} mappings. The resulting output sequences of the attention heads are combined (e.g. concatenated, or discarded except for one classification token in classification tasks as ours) and presented to a final layer that transforms the latent vectors to a specific output, e.g. a softmax over a vocabulary (in language tasks), or a softmax over classes, such as the bat species in our case. For the sake of brevity, we have skipped some details of this architecture, such as residual connections, layer normalization, etc., for which we refer the reader to the original paper \cite{AIAYN}.

\begin{figure}
    \centering
    \includegraphics[width=\textwidth]{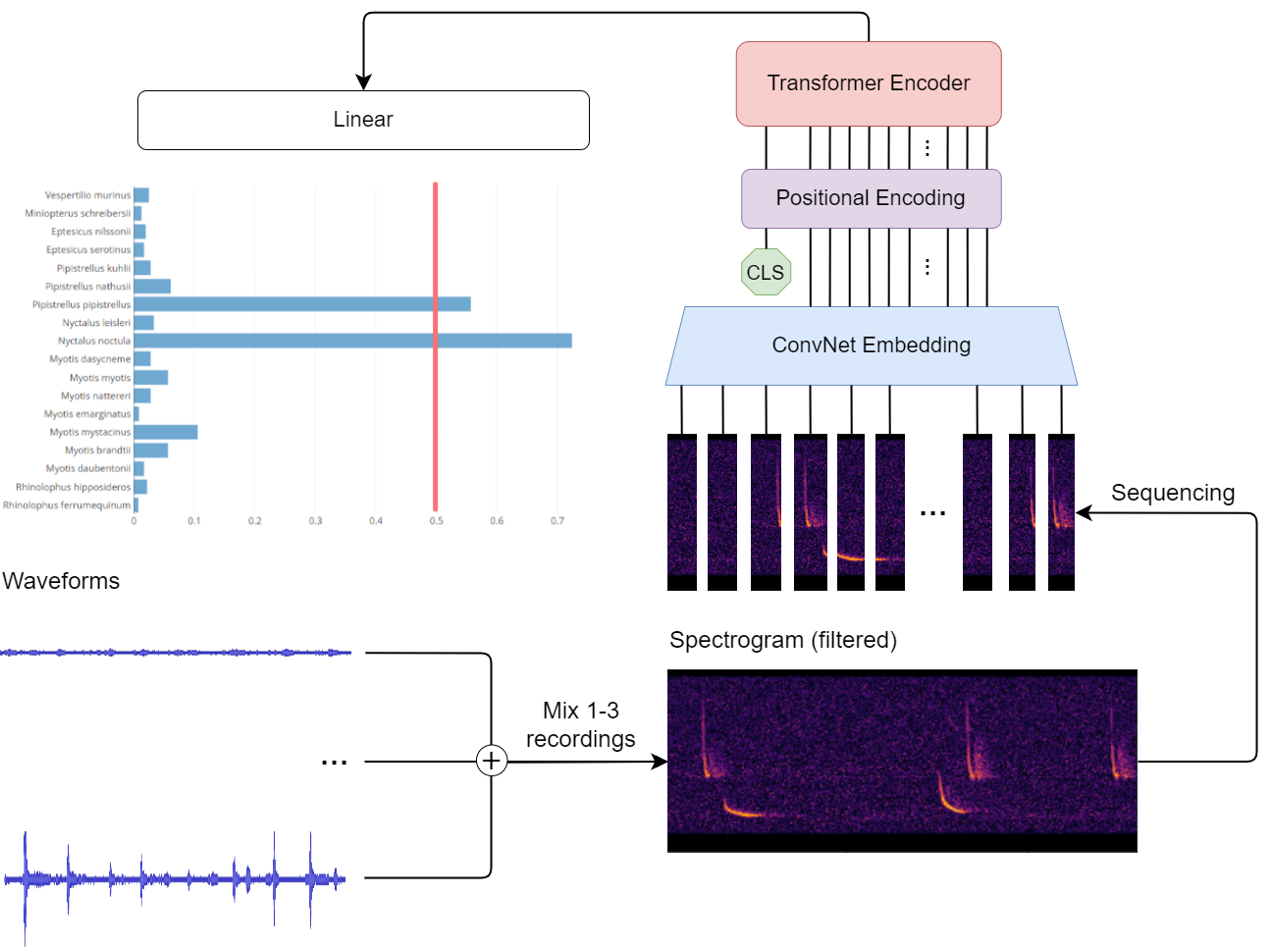}
    \caption{The proposed model architecture.}
    \label{fig:model}
\end{figure}

As already mentioned, the input to our transformer network is provided by ConvNet patches. To obtain these patches, each sample from our preprocessed dataset of mixed call sequences is sliced into a sequence of 60 overlapping patches (with 50\% overlap). Each of the resulting patches is embedded using the same ConvNet consisting of three blocks of convolution, batch normalization, ReLu activation and max pooling. The embedding size for each patch is 64. Similar to other Transformer-type classification networks, such as BERT \cite{BERT}, a classification (CLS) token is appended to the token sequence. The resulting sequence of patch embeddings and CLS token is then fed into a small Transformer-type encoder consisting of two self-attention layers with two attention heads each, and a feed-forward dimension of 32. A final linear layer and sigmoid activation applied to the transformed CLS token produces the output of the network, predicting the detected bat calls. The model is trained on mixed sequences and multiple labels for multi-label classification, where a class is considered positive whenever the sigmoid of the logits for that class is above 0.5. We manually optimized our model using the validation set. The best results were obtained using Asymmetric Loss \cite{ASL}, Sharpness-Aware Minimization \cite{SAM}, cosine scheduler and learning rate of 5e-4 with 25 epochs. Compare Figure \ref{fig:model} for a visualization of the model architecture.\\



\section{Results}
\label{sec:results}
Evaluation was conducted on two types of samples: single species, where only one species is present in a sample, and mixed species, where multiple species are present in a sample. The mixed species samples were synthetically generated using the method described in Section \ref{sec:dataprep}). By conducting these comparative analyses, we gained insights into the model's performance on single species samples as well as its adaptability to mixed species sequences. Two metrics were utilized for evaluation: accuracy and F1-score. Accuracy was employed in the evaluation of single species samples, as it measures the proportion of correctly classified instances. However, for multi-label classification, accuracy is not defined, and therefore, it was used exclusively in the single species evaluation.
In contrast, the F1-score was employed in both single species and mixed species evaluations. It combines precision and recall into a single measure, considering both true positives and false negatives in the dataset. This metric proves particularly effective in situations where the dataset is unbalanced, enabling a comprehensive evaluation of the model's performance \cite{Han2023}."

\subsection{Single species}
To establish a baseline, we initially assess the performance solely on single species samples before proceeding to evaluate sequences with mixed species in the subsequent section. To this end, we replicate the setting of Schwab et al. \cite{schwab}, which utilizes a ResNet architecture. Their model operates on individual calls extracted from call sequences using peak detection and a secondary ResNet. In addition to replicating their approach, we further explored the baseline model's capabilities by incorporating sequences of individual calls and averaging the predictions (referred to as Baseline sequential). This allowed us to assess the model's performance when presented with sequential call data.

\begin{figure}
    \centering
        \begin{tabular}[H]
        { |p{5cm}||p{2cm}|p{2cm}|p{2cm}|  }
        \hline
        \multicolumn{4}{|c|}{Trained on single species} \\
        \hline
        Model & Accuracy & F1-score & \# Params \\
        \hline
        Baseline & 76.53\% & 68.37\% & 6,148,563 \\
        Baseline sequential & 78.49\% & 68.96\% & 6,148,563 \\
        BAT ResNet & \textbf{83.35\%} & \textbf{81.51\%} & 6,383,634\\
        BAT & 82.15\% & 78.42\% & 69,970\\
        MLP & 79,20\% & 74,41\% & 22,198,034\\
        \hline
        \multicolumn{4}{|c|}{Trained on mixed species} \\
        \hline
        BAT & 82.18\% & 77.98\% & 69,970\\
        \textcolor{gray}{BAT no val} & \textcolor{gray}{88.92\%} & \textcolor{gray}{84.23\%} & \textcolor{gray}{69,970}\\
        \hline
        \end{tabular}
    \caption{Comparison of different architectures on single species recordings.}
    \label{tbl:single}
\end{figure}

As one can see from Figure \ref{tbl:single}, our Transformer-based approach shows better test performance than the baseline regarding accuracy and F1-score, both when trained on single or mixed species recordings. In addition to our regular model (BAT), Figure \ref{tbl:single} also shows the performance of two variations, where in one case we replace the ConvNet with a larger ResNet, and in the other case we replace the Transformer with a two-layer MLP. Both variations consist of significantly more parameters than BAT. The MLP version showed notably worse performance, whereas the ResNet version was slightly better, but required about 100 times the parameters. This shows that the combination of ConvNet embedding and Transformer is well-suited for the task. 
We also checked for potential improvements if more data were available, bu adding the validation set to the training dataset---this is indicated by the gray results in Figure \ref{tbl:single}. This type of improvement is of course not expected to be special for our model.

\subsection{Mixed species}

\begin{figure}[H]
    \centering
        \begin{tabular}[H]
        { |p{3cm}||p{2.5cm}|p{2.5cm}|p{2.5cm}|  }
        \hline
        \multicolumn{4}{|c|}{Trained on single species} \\
        \hline
        Model & Micro F1 & Macro F1 & \# Params \\
        \hline
        Baseline sequential & 30.85\% & 27.93\% & 6,148,563 \\
        BAT & 46.57\% & 40.27\% & 69,970\\
        \hline
        \multicolumn{4}{|c|}{Trained on mixed species} \\
        \hline
        Baseline sequential & 64.15\% & 50.00\% & 6,148,563 \\
        MLP & 72.21\% & 60.89\% & 22,198,034 \\
        LSTM & 76.82\% & 68.85\% & 94,866 \\
        ConvNet & \textbf{77.4\%} & 69.12\% & 124,162 \\
        Small ConvNet & 74.09\% & 63.72\% & \textbf{46,114} \\
        BAT & 76.62\% & \textbf{69.31\%} & 69,970\\
        \textcolor{gray}{BAT no val} & \textcolor{gray}{83.02\%} & \textcolor{gray}{77.17\%} & \textcolor{gray}{69,970} \\
        \hline
        \end{tabular}
    \caption{Comparison of different architectures on multi-label classification of mixed species recordings. The top two models were trained on single species data, whereas the models in the bottom were trained on mixed species data.}
    \label{tbl:mixed}
\end{figure}

When testing predictions on mixed species recordings (see Figure \ref{tbl:mixed}), our Transformer-based model BAT significantly outperforms the baseline model, both when trained on single species and mixed species recordings. Overall, BAT performs similar to other state of the art models such as LSTMs and ConvNets. While it performs roughly the same as an LSTM with approximately $95,000$ parameters, BAT requires less than $70,000$ parameters . BAT's performance sits in between a small and large ConvNet that we tested, with about 65\% and 175\% the number of parameters, respectively. However ConvNets lack the ability of variable input lengths, that is, the size of the input image must be predefined and consistent for all images in the dataset. For Transformers, the sequence length can be increased and shorter inputs can just be padded.

\subsection{Comparison to available software}
\label{stc:comparison}
In this section, we compare our method to other, mostly commercially available software, like BatExplorer, batIdent and bdAnalyzer  \cite{schwab}. For comparison, we used 704 samples from our test set selected from the Skiba dataset \cite{skiba2003europäische}, where each recording lasts 780 ms, and 167 samples from another smaller bat call dataset called ChiroVox \cite{Grfl2022}, where each recording lasts between 1-10 seconds. To make the comparison fairer,
we only used species that both BAT and bdAnalyzer did train on. If 0 calls were detected and thus no classification can be made, the sample classification was counted as incorrect. Our model and bdAnalyzer are biased towards the Skiba dataset because both trained on parts of it. The ChiroVox \cite{Grfl2022} dataset is completely independent. BatExplorer \cite{BATLOGGE94:online} could only export two detected species, so all mixed sequences with more than 2 were removed when testing BatExplorer. 
We used default settings for all tools, for bdAnalyzer on the Skiba dataset we used a manual call detection threshold of 0.3 instead of the automatic threshold, because otherwise too few calls were detected. From our validation set we could deduce, that a multi-label prediction threshold of 0.33 yields the best results for our model. For all other methods a threshold of 0.5 was used.

\begin{figure}[H]
    \centering
        \begin{tabular}[H]
        { |p{5cm}||p{2cm}|p{2cm}|p{2cm}|  }
        \hline
        \multicolumn{4}{|c|}{Skiba - Single species} \\
        \hline
        Model & Accuracy & Micro F1 & Macro F1 \\
        \hline
        batIdent & 22.8\% & 35.34\% & 21.62\% \\
        BatExplorer & 38.15\% & 46.48\% & 34.36\% \\
        bdAnalyzer & 64.13\% & 71.71\% & 60.56\% \\
        BAT & \textbf{84.19\%} & \textbf{84.58\%} & \textbf{79.52\%} \\
        \hline
        \multicolumn{4}{|c|}{ChiroVox - Single species} \\
        \hline
        batIdent & 24.03\% & 38.51\% & 12.52\% \\
        BatExplorer & 16.28\% & 25.15\% & 10.15\% \\
        bdAnalyzer & 46.27\% & 56.11\% & 24.51\% \\
        BAT & \textbf{72.09\%} & \textbf{77.18\%} & \textbf{51.05\%} \\
        \hline
        \end{tabular}
    \caption{Comparison of different commercially available tools for classification on single species recordings from Skiba \cite{skiba2003europäische} and ChiroVox \cite{Grfl2022} database.}
    \label{tbl:comp_single}
\end{figure}

\begin{figure}[H]
    \centering
        \begin{tabular}[H]
        { |p{5cm}||p{2cm}|p{2cm}|  }
        \hline
        \multicolumn{3}{|c|}{Skiba - Mixed species} \\
        \hline
        Model & Micro F1 & Macro F1 \\
        \hline
        batIdent & 22.48\% & 14.08\% \\
        BatExplorer & 41.84\% & 33.18\% \\
        bdAnalyzer & 65.56\% & 57.93\% \\
        BAT & \textbf{75.89\%} & \textbf{70.42\%} \\
        \hline
        \multicolumn{3}{|c|}{ChiroVox - Mixed species} \\
        \hline
        batIdent & 45.14\% & 12.67\% \\
        BatExplorer & 50.51\% & 22.30\% \\
        bdAnalyzer & 52.13\% & 29.42\% \\
        BAT & \textbf{69.91\%} & \textbf{36.32\%} \\
        \hline
        \end{tabular}
    \caption{Comparison of different commercially available tools for classification on mixed recordings from Skiba \cite{skiba2003europäische} and ChiroVox \cite{Grfl2022} database.}
    \label{tbl:comp_mixed}
\end{figure}

Our method outperforms every commercially available tool, and that at a smaller computational footprint than all the other methods, opening up the possibility for real-time deployment and real-time species classification.


\section{Discussion}

Our study demonstrates the potential applicability of Transformer-based models for efficient classification of bioacoustic signals, such as bat call classification, allowing for high quality real-time detection based on a light-weight model.
Most previous methods for bat call classification were trained on short recordings consisting of single bat calls \cite{schwab, tabak, Dierckx}, which can make identification much more difficult compared to longer recordings with multiple calls \cite{Aodha2022}. However, longer recordings come with their own difficulties, including the necessity for larger models and larger variability of the data. In fact, the presence of multiple species calls in longer recordings has been previously pointed out as one of the main challenges in bat detection \cite{Dierckx, Aodha2022}.

\begin{figure}[H]
    \centering
    \begin{tabular}{l|c|c|c|c|c}
    Model & Detect & Classify & \makecell{Call\\ sequence} & \makecell{Multi-\\species} & \makecell{Simple\\ annotation} \\
    \hline
    Bat detective \cite{batdetective} & Yes & No & No & No & No\\ 
    Schwab et al. \cite{schwab} & No & Yes & No & No & Yes \\ 
    Tabak et al. \cite{tabak} & No & Yes & No & No & Yes \\
    Zualkernan et al. \cite{Zualkernan} & No & Yes & No & No & Yes \\
    Chen et al. \cite{CHEN} & No & Yes & No & No & Yes \\
    Dierckx et al. \cite{Dierckx} & No & Yes & No & Yes & Yes \\
    Alipek et al. \cite{Alipek} & No & Yes & Yes & No & Yes \\
    Batdetect2 \cite{Aodha2022} & Yes & Yes & Yes & (Yes) & No \\ 
    Ours & Yes & Yes & Yes & Yes & Yes \\
    \end{tabular}
    \caption{Comparison of multiple related works and their characteristics.}
    \label{tbl:related_models}
\end{figure}

Previously, multi-label classification of non-overlapping calls was only possible by classifying each call individually in a sequence, leaving out temporal information of call sequences \cite{schwab, Dierckx}. A comparison of different models can be seen in Figure \ref{tbl:related_models}. Here, we compare different model characteristics, for example whether the model is able to detect individual calls or whether the model is capable of species classification. Most models only focus on species identification \cite{schwab, tabak, Zualkernan, CHEN, Dierckx, Alipek}, without predicting specific call locations \cite{batdetective, Aodha2022}. Interestingly, our model is able to predict call locations indirectly as a side effect of creating patches and leveraging the attention mechanism of the Transformer. Another characteristic we compare, is whether the model is able to use temporal information from sequences of calls, where most models only aim to detect or identify single calls and only two make predictions on sequences of calls \cite{Alipek, Aodha2022}. Most models were trained on single-species recordings and thus are not capable of detecting overlapping calls. Only a few models implemented a multi-label approach \cite{Dierckx, Aodha2022}. In particular, Batdetect2 \cite{Aodha2022} follows an exceptional approach, where multi-label classification is used, but the detection of overlapping calls is suppressed through Non-Maximum-Suppression. Models that are trained on individual calls are inherently capable of multi-species classification within a sequence of non-overlapping calls by classifying each call separately, with the downside of disregarding temporal information. The last characteristic we compare is whether the model is trained on data that was extensively annotated. In Batdetective \cite{batdetective}, this involved annotating each call individually with bounding boxes, whereas in Batdetect2 \cite{Aodha2022}, not only bounding boxes but also class labels were annotated. Obtaining the necessary resources for such costly annotations remains a challenge in acoustic monitoring in most places, thereby limiting its adoption.

Importantly, in our study, we do not learn features solely from single calls, but our model is trained on a synthetically created dataset of multi-species recordings, and thus can make use of temporal information and changes between calls. Randomly mixing the recordings might also have served as augmentation, resulting in more robust latent representations. Our results show that the combination of the ConvNet and Transformer architecture borrowed from computer vision \cite{AIAYN, levit} provides an efficient model with a moderate number of parameters that can successfully cope with this increased variability of the data. This allows our model to improve on most challenges of previous models that we mentioned in the introduction \cite{schwab, tabak}, such as the difficulty of overlapping calls, despite being light-weight and therefore easy to train and deploy.

Although the Transformer-ConvNet architecture is in principle a black-box model, the attention mechanism allows to highlight relevant calls for species identification through attention maps \cite{An} (compare Appendix \ref{fig:cam_out}). The self-attention mechanism has recently been reported to improve bat call classification in another study \cite{Aodha2022} that segmented multi-call recordings for multi-species classification. Aoadha et al. used a dataset of short bat call recordings that were annotated by bounding boxes and class labels of individual calls. This costly designed dataset was then leveraged to train an encoder-decoder ConvNet with an attention mechanism on the latent space to predict bounding boxes and species of individual calls. In contrast, our approach uses a much simpler dataset and longer sequences, while still being able to visualize the most informative calls for a specific prediction through attention maps. Their model achieves similar performance on a much larger dataset with comparable classes to our Skiba dataset. Additionally, Aodha et al. excluded acoustic behaviours such as feeding buzzes and social calls from their dataset. We, on the other hand, intentionally incorporated them to make species predictions on those particular acoustic behaviours.

While our model provides a first step towards direct multi-species classification, there is considerable room for improvement. Particularly, for mixed-species training the main challenge is posed by limited annotated data and imbalanced species occurrence. The training dataset of multi-species call sequences that we artificially created from single-species recordings ideally should be replaced by a dataset of actual recordings of mixed species, which might differ quite a bit from simply adding signals. Also, training on more diverse data from multiple different datasets would benefit generalization to unseen data \cite{Aodha2022}. Moreover, we were able to significantly increase the classification performance by including the validation dataset into the training data, reflecting the fact the limitation in data availability might actually be the culprit of current model performance. In fact, since classifying bat calls needs expert knowledge and takes a lot of time, there is very little annotated data. Additionally, the occurrence of different species is highly unbalanced, which is reflected in currently available datasets. One possibility to deal with this issue could be the inclusion of unsupervised training.


\section{Conclusion}

In this work, we propose a new model for bat call classification. We use a ConvNet-Transformer hybrid model to classify sequences of bat calls, instead of only classifying single bat calls as in previous approaches. Additionally, by synthesizing mixed call sequences out of single call sequences, we were able to incorporate multi-label classification for classifying call sequences where multiple species are calling at the same time. Without using multi-stage classification models, we found new state-of-the-art results, that even outperform commercially available tools and other methods (Section \ref{stc:comparison}). In particular, we were able to achieve a single species accuracy of 88.92\% (F1-score of 84.23\%) and a multi species macro F1-score of 74.40\% on our test set. On another, independent dataset we achieved a single species accuracy of 72.09\% (F1-score of 51.05\%) and a multi-species macro F1-score of 36.32\%.

As a final remark, we want to note that our model is not tuned in any way for bats specifically. Hence, the same architecture could also be applied to other domains like bird or whale call classification, where a light-weight model like ours might have similar advantages over other approaches.



\printbibliography

\section*{Appendix}

\subsection{Availability}
We provide a demo web implementation for the trained model available at \url{https://bat.hadros.de/}.
The user is provided with two options: they can either select from a set of example files or upload their own WAV file. If the recording has already been time expanded by 1:10, the user must specify this. The selected audio file is displayed in a minimalistic wave format, allowing playback functionality.

Upon selecting a desired model and clicking the 'predict' button, the audio is sent to the server. There, the audio undergoes pre-processing, is divided into overlapping patches, and is fed through the chosen model. In addition to the model's output, a Grad-CAM visualization is generated for each predicted label and sent back to the client. This visualization includes the original spectrogram, activation maps, and the prediction.

The predicted labels are displayed as tabs, and clicking on a specific tab reveals the corresponding activation map. Due to memory restrictions in this demo, only the first 60 patches (780 ms) are utilized. It is worth noting that the web demo may exhibit slower performance due to data transfer to and from the server, but the inference process itself is fast. For real-world applications, it is recommended to use the model offline. To facilitate this, we have developed a command-line tool available on GitHub (\url{https://github.com/FrankFundel/BAT-cli}). The tool can be cloned from the repository, and its usage is straightforward, with comprehensive documentation provided on the GitHub page. By passing a directory as an argument to the CLI, all files within that directory will be processed and classified, with the results conveniently saved in a CSV file.

\subsection{Details about the preprocessing of our data} \label{appendix:preprocessing}
We created two functions \emph{getIndividuals} which extracts individual calls from the recordings and \emph{getSequences} which extracts patch-sequences from the recordings.

In \emph{getIndividuals}, sound events are detected and, if classified as bat calls, these sound events were cut out surrounded by a window-patch of a certain size. Since n\_fft=512, 23 ms of audio (230 ms time expanded) resulted in 512 samples and the average call length is ca. 25 ms (250 ms time expanded, calculated using data from Skiba \cite{skiba2003europäische}), an appropriate patch length is 44 with an overlap of 22. To detect sound events, the mean over each time step was calculated and the built-in function for peak detection from the python audio processing library librosa \cite{Librosa90:online} was used. To differentiate between noise and an actual call, we set up a small ResNet-18 to classify between those two classes and only return patches that were classified as a bat call (inspired by \cite{schwab}). For that we manually classified over 2,400 patches as call/no-call and achieved a test accuracy of 94.77\% (ADAM, ReduceLROnPlateau, 0.001 initial learning rate, batch size 128 for 35 epochs). The \emph{getIndividuals} function returns 33,978 labeled and classified call patches.

The \emph{getSequences} slices the whole spectrogram into patches of size 44, and then slices the consecutive patches again, resulting in overlapping sequences of overlapping patches. Since the average calls per second is around 9 (calculated using data from Skiba \cite{skiba2003europäische}), we selected a sequence length of 60 patches (1 second) and a sequence overlap of 15 patches (250 ms). No peak detection or call/no-call classification is needed, since empty patches are important for the preservation of time information.

\subsection{Visualization of the attention mechanism in our model}
We can use Grad-CAM \cite{Selvaraju_2019} to visualize the activation of the ConvNet and the attention of the Transformer part of our final model (mixed BAT ConvNet). Grad-CAM uses the gradients of a specific label during inference, to create a heatmap of the most "important" parts of an any input with respect to this target. First a target layer or multiple target layers needs to be specified, then Grad-CAM will follow the gradients that flow into this layer to calculate the activation map. Usually this is some kind of normalization layer, so we chose the first normalization layer of the ConvNet and the first normalization layer of the Transformer. A custom reshape method is passed to the Grad-CAM algorithm, that transforms the input with respect to each target layer so that the result can be displayed as an image. The activation maps for each target layer are summed up to create a final activation map. The predicted labels of the model are then used to create separate activation maps for each individual label. The activation map is multiplied element-wise with the original input sequence to create a masked output sequence. A few examples are shown in the appendix (Fig. \ref{fig:cam_out}).

\begin{figure}[H]
    \centering
    \caption{Ground truth with input sequence is on top, followed by masked input sequence for each predicted label.}
    \label{fig:cam_out}
    \includegraphics[width=\textwidth]{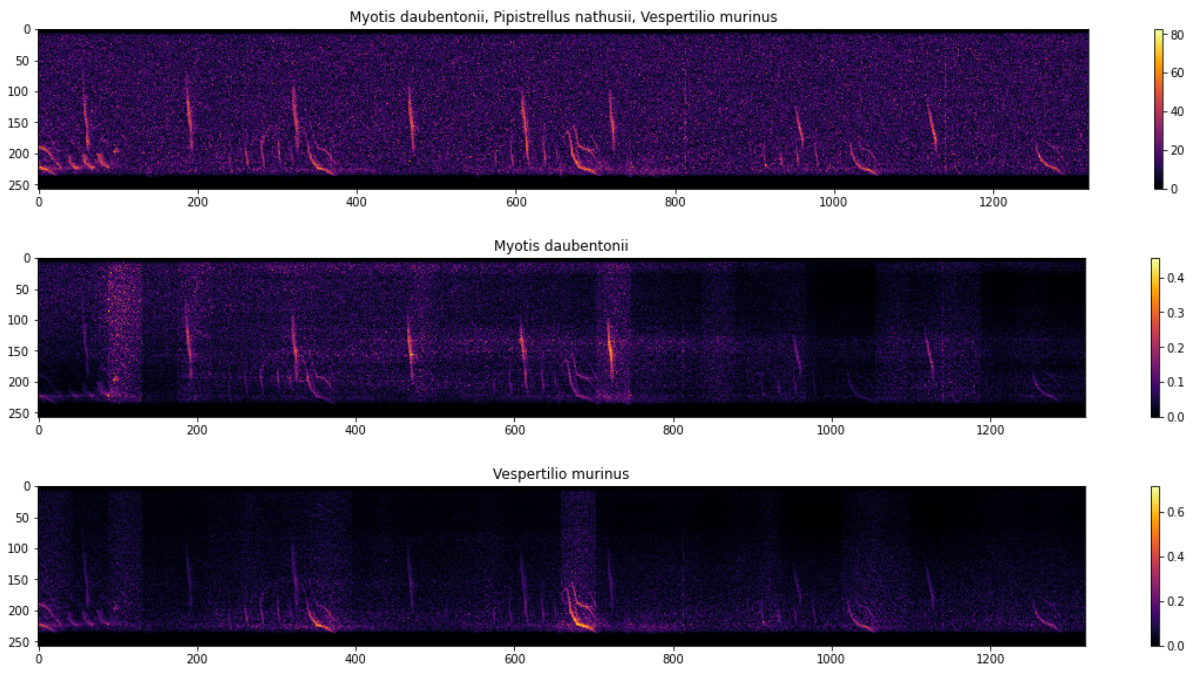}
\end{figure}
\hrule
\begin{figure}[H]
    \centering
    \includegraphics[width=\textwidth]{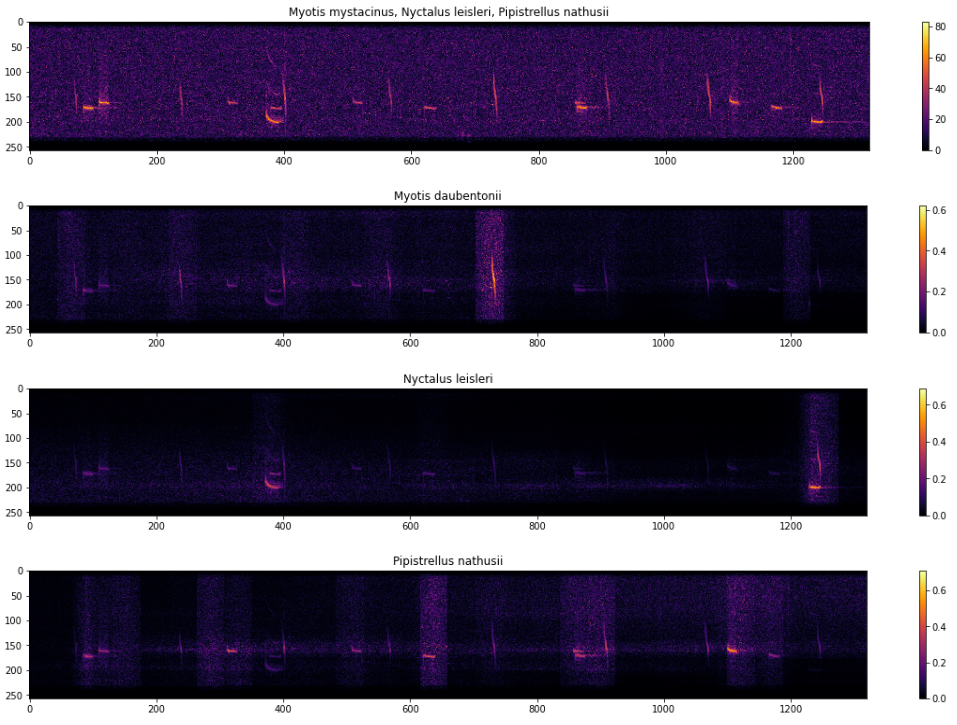}
\end{figure}
\hrule
\begin{figure}[H]
    \centering
    \includegraphics[width=\textwidth]{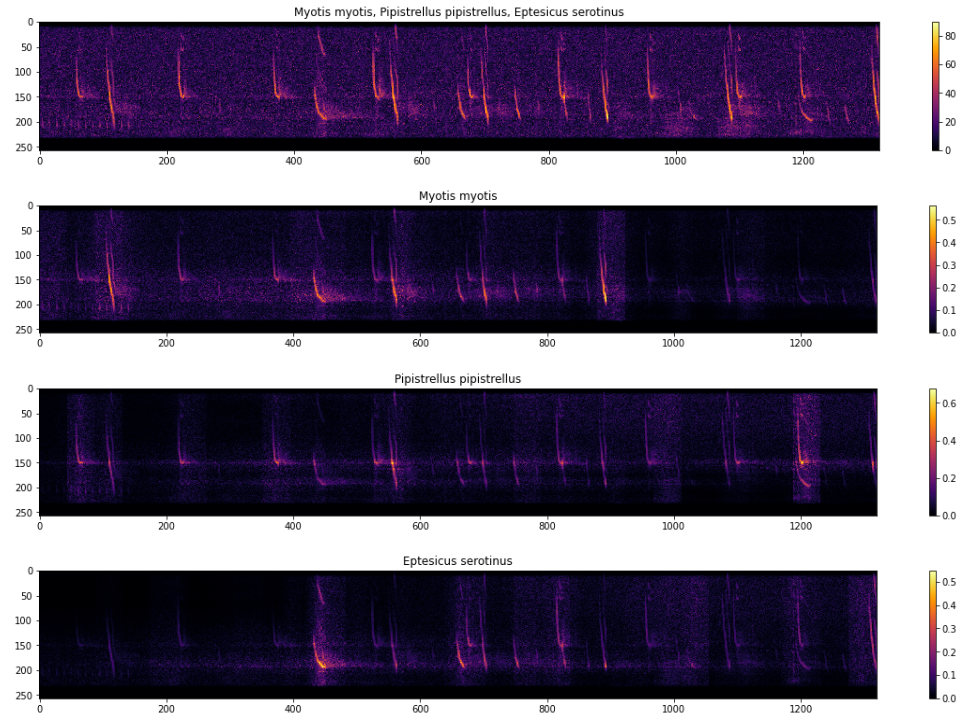}
\end{figure}
\hrule
\begin{figure}[H]
    \centering
    \includegraphics[width=\textwidth]{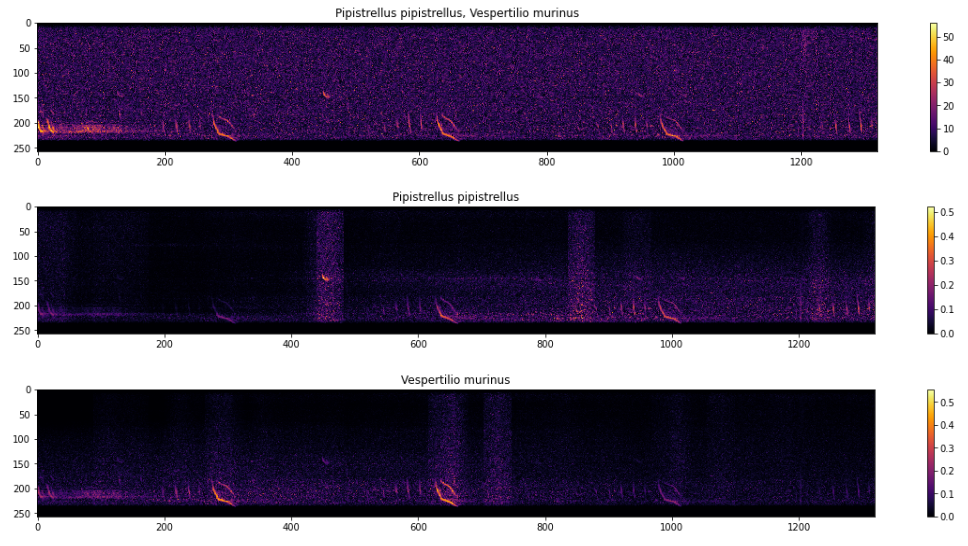}
\end{figure}
\hrule
\begin{figure}[H]
    \centering
    \includegraphics[width=\textwidth]{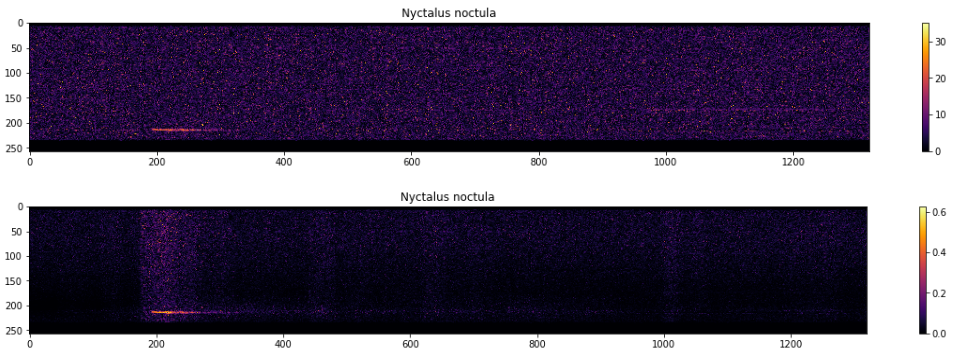}
\end{figure}
\hrule
\begin{figure}[H]
    \centering
    \includegraphics[width=\textwidth]{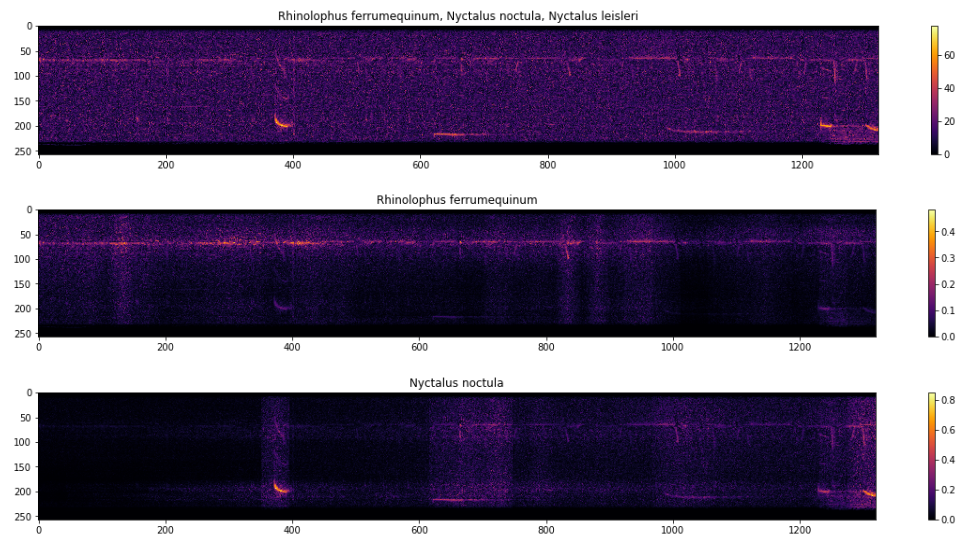}
\end{figure}
\hrule
\begin{figure}[H]
    \centering
    \includegraphics[width=\textwidth]{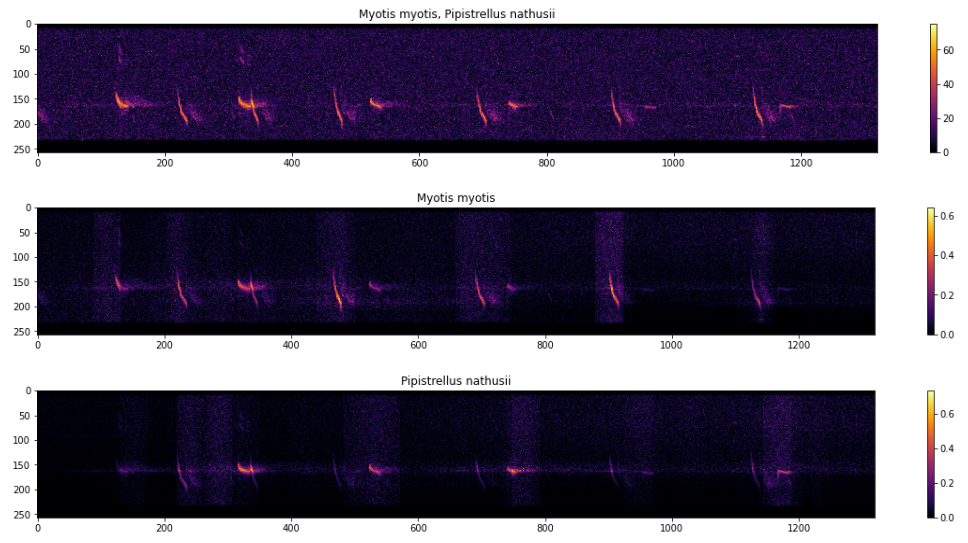}
\end{figure}
\hrule
\begin{figure}[H]
    \centering
    \includegraphics[width=\textwidth]{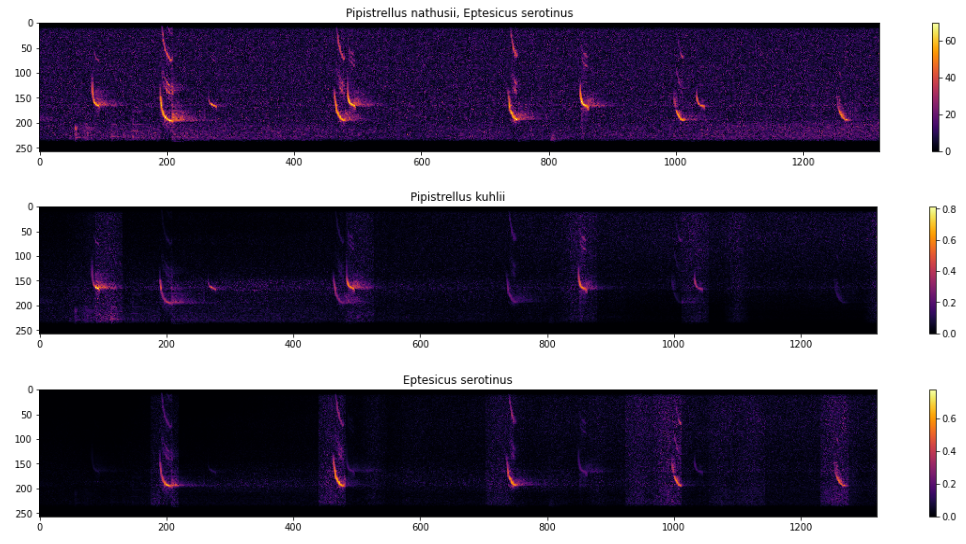}
\end{figure}



\end{document}